\documentclass[10pt,twocolumn,letterpaper]{article}

\usepackage{silence}
\usepackage{amsmath,amsfonts,bm}

\def\eqref#1{equation~\ref{#1}}

\def\1{\bm{1}}

\DeclareMathAlphabet{\mathsfit}{\encodingdefault}{\sfdefault}{m}{sl}
\SetMathAlphabet{\mathsfit}{bold}{\encodingdefault}{\sfdefault}{bx}{n}

\usepackage{iccv}
\usepackage{times}
\usepackage{epsfig}
\usepackage[pagebackref=true,breaklinks=true,letterpaper=true,colorlinks,bookmarks=false]{hyperref}

\usepackage{url}
\usepackage{booktabs}       %
\usepackage[utf8]{inputenc} %
\usepackage[T1]{fontenc}    %
\usepackage{amsfonts}       %
\usepackage{nicefrac}       %
\usepackage{microtype}      %
\usepackage{color}
\usepackage{dsfont}
\usepackage{setspace}
\usepackage{amsmath,amssymb}
\usepackage{rotating}
\usepackage{multirow}
\usepackage{xspace}
\usepackage{tabularx,ragged2e}
\usepackage{array}
\usepackage{indentfirst}
\usepackage{float}
\usepackage{adjustbox}
\usepackage{graphicx}
\usepackage{subfigure}
\usepackage[font=small]{caption}
\usepackage{makecell}
\usepackage{arydshln}
\usepackage{hhline}
\usepackage{algorithm,algpseudocode}
\usepackage{algorithmicx}
\usepackage[accsupp]{axessibility}
\usepackage{tabularray}

\newcolumntype{P}[1]{>{\centering\arraybackslash}p{#1}}
\newcolumntype{M}[1]{>{\centering\arraybackslash}m{#1}}

\newcommand{\myparagraph}[1]{\vspace{-1pt}\paragraph{#1}}
\newcommand{\ours}{DenseDiffusion\xspace}

\iccvfinalcopy %

\def\iccvPaperID{6848} %

\ificcvfinal\fi

\begin{document}

\title{Dense Text-to-Image Generation with Attention Modulation}

\author{Yunji Kim$^1$ \qquad Jiyoung Lee$^1$ \qquad Jin-Hwa Kim$^1$ \qquad Jung-Woo Ha$^1$ \qquad Jun-Yan Zhu$^2$\\ \\
$^1$NAVER AI Lab \qquad $^2$Carnegie Mellon University\\
}

\ificcvfinal\thispagestyle{empty}\fi

\twocolumn[{
\maketitle
\centering
    \begin{tabular}{ccc}
     &\multicolumn{1}{c}{\qquad \qquad \qquad \qquad \qquad \qquad \quad \small{DenseDiffusion (Ours)}}
     &\multicolumn{1}{c}{\qquad \qquad \qquad \quad \small{Stable Diffusion~\cite{ldm}}} \\
    \multicolumn{3}{c}{\includegraphics[width=0.97\linewidth]{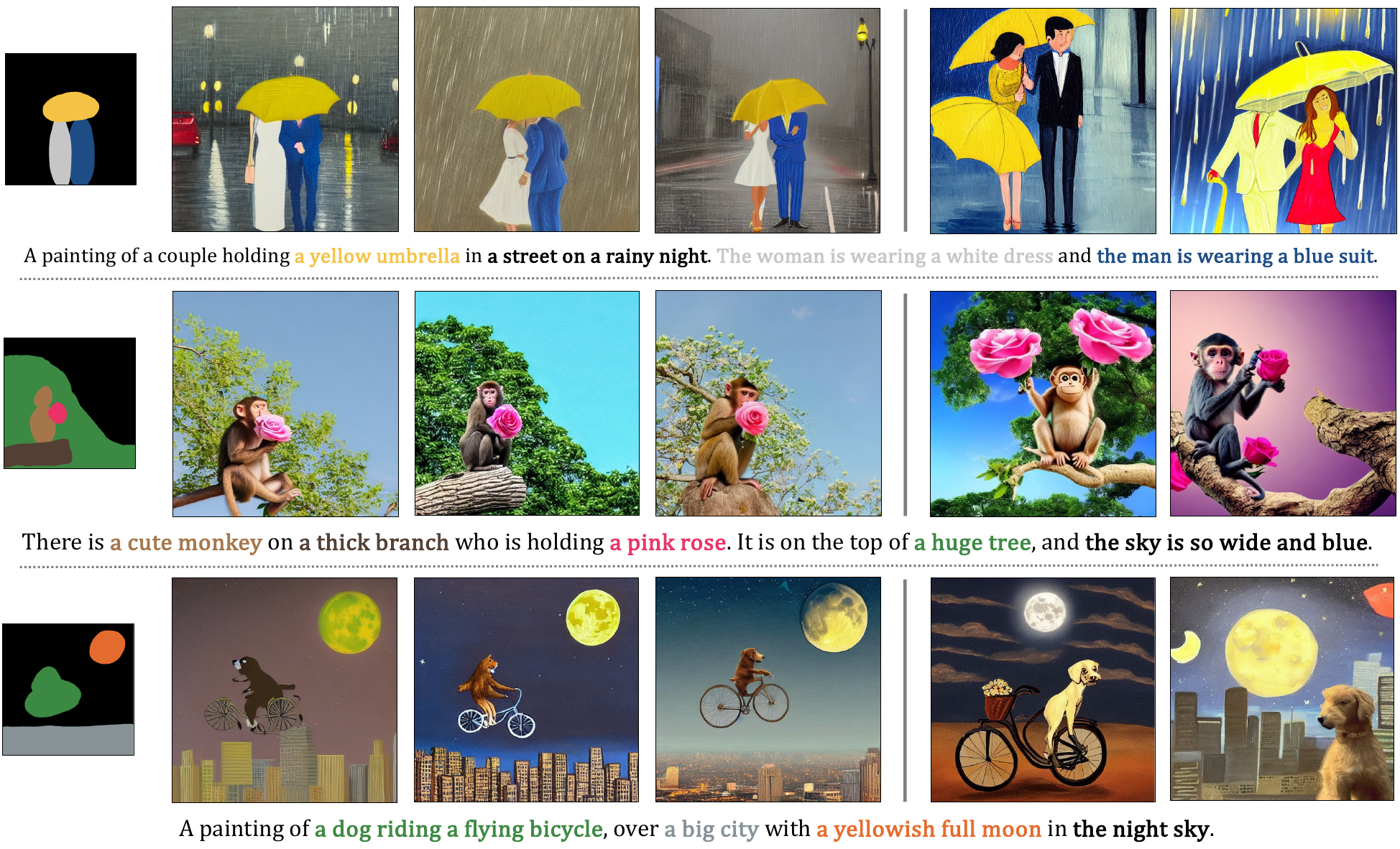}} \\
    \end{tabular}
  \captionof{figure}{
DenseDiffusion enables us to incorporate both text and layout information into pre-trained text-to-image models without requiring additional fine-tuning. Our method not only synthesizes images that follow text prompts more faithfully but also offers users greater control over object and scene layout. We achieve it by modulating attention maps of pre-trained models, such as Stable Diffusion~\cite{ldm}, according to both text and layout conditions. 
  }

  \label{fig:teaser}
 \vspace*{0.6cm}
}]

\begin{abstract}
\vspace*{-0.07 cm}
Existing text-to-image diffusion models struggle to synthesize realistic images given dense captions, where each text prompt provides a detailed description for a specific image region.
To address this, we propose DenseDiffusion, a training-free method that adapts a pre-trained text-to-image model to handle such dense captions while offering control over the scene layout.
We first analyze the relationship between generated images' layouts and the pre-trained model's intermediate attention maps.
Next,  we develop an attention modulation method that guides objects to appear in specific regions according to layout guidance.
Without requiring additional fine-tuning or datasets, we improve image generation performance given dense captions regarding both automatic and human evaluation scores.
In addition, we achieve similar-quality visual results with models specifically trained with layout conditions.
Code and data are available at \url{https://github.com/naver-ai/DenseDiffusion}.
\end{abstract}
\vspace{-20pt}
\section{Introduction}
\label{sec:introduction}

Recently developed text-to-image models can synthesize diverse, high-quality images from short scene descriptions~\cite{ddpm, guided_diffusion, glide, dalle2, imagen, ldm,kang2023scaling,yu2022scaling,chang2023muse}.
However, as shown in Figure~\ref{fig:teaser}, they often encounter challenges when handling dense captions and tend to omit or blend the visual features of different objects. 
Here the term ``dense'' is inspired by the work of dense captioning~\cite{densecap} in which each phrase is used to describe a specific image region.
Moreover, it is difficult for users to precisely control the scene layout of a generated image with text prompts alone. %

Several recent works propose providing users with spatial control by training or fine-tuning layout-conditioned text-to-image models~\cite{spatext, sketchguide, controlnet,gligen, ldm, makeascene,wang2022pretraining,mou2023t2i}. While Make-a-Scene~\cite{makeascene} and Latent Diffusion Models~\cite{ldm} train models from scratch given both text and layout conditions, other recent and concurrent works, such as SpaText~\cite{spatext} and ControlNet~\cite{controlnet}, introduce additional spatial controls to existing text-to-image models with model fine-tuning. Unfortunately, training or fine-tuning a model can be computationally-expensive. Moreover, the model needs to be retrained for every new user condition, every new domain, and every new text-to-image base model.

To address the above issue, we propose a training-free method that supports dense captions and offers layout control. 
We first analyze the intermediate features of a pre-trained text-to-image diffusion model to show that the layout of a generated image is significantly related to self-attention and cross-attention maps.
Based on this observation, we modulate intermediate attention maps according to the layout condition on the fly. 
We further propose considering the value range of original attention scores and adjusting the degree of modulation according to the area of each segment.

Our experiments show that our method improves the performance of Stable Diffusion~\cite{ldm} and outperforms several compositional diffusion models~\cite{compdiff,multidiffusion,strucdiff} on dense captions regarding both text conditions, layout conditions, and image quality. We verify them using both automatic metrics and user studies. %
In addition, our method is on par with existing layout-conditioned models regarding qualitative results. Finally, we include a detailed ablation study regarding our design choices and discuss several failure cases. Our code, models, and data are available at \url{https://github.com/naver-ai/DenseDiffusion}.

\section{Related work}
\label{sec:related_work}

\myparagraph{Text-to-Image Diffusion Models.}

The development of large-scale diffusion models~\cite{glide, dalle2, imagen, ldm} has enabled us to synthesize diverse photorealistic images with flexible editing capacity~\cite{sdedit,choi2021ilvr,kim2022diffusionclip,blended_latent,kawar2023imagic}. 
In particular, recent works~\cite{prompt_to_prompt, plugnplay, diffedit,parmar2023zero} edit images by analyzing the interaction between intermediate image features and input textual features during the generation process.
Prompt-to-Prompt~\cite{prompt_to_prompt} extracts cross-attention maps for each text token and injects them when denoising with an edited caption to preserve the input structure. 
Plug-and-Play~\cite{plugnplay} takes a similar approach but
reuses self-attention and spatial features. 
pix2pix-zero~\cite{parmar2023zero} proposes cross-attention guidance to encourage the cross-attention maps of original and edited images to stay close. %
Inspired by the above works, we seek to leverage the close connection between the text features and the image layout inside a pre-trained diffusion model but focus on image synthesis with spatial controls rather than image editing. %

\myparagraph{Compositional Diffusion Methods.}

Since text-to-image models are typically trained on datasets with short text captions, %
 they often fail to reflect all the details of dense captions composed of several phrases.
A line of work tries to address the problem with a pre-trained model in a training-free manner. %
For example, Composable Diffusion~\cite{compdiff} and MultiDiffusion~\cite{multidiffusion} perform a separate denoising process for each phrase at every timestep. Attend-and-Excite~\cite{attendnexcite} optimizes the noise map to increase the activation of the neglected tokens in cross-attention maps. %
Structure Diffusion~\cite{strucdiff} indirectly modulates cross-attention scores by enriching textual features for parsed texts. %

Meanwhile, similar to the work of Dong et al.~\cite{onthefly} that directly increases attention scores of certain words to increase the likelihood of them being generated for language modeling tasks, Paint-with-words (eDiffi-Pww)~\cite{ediffi} increases the cross-attention scores between image and text tokens that describe the same object given segmentation masks.
However, they sometimes fail to reflect text and layout conditions faithfully.
In this work, we propose a more extensive modulation method by devising multiple regularization terms.

\myparagraph{Image Synthesis with Spatial Control.} 
Seminal works~\cite{chen2009sketch2photo,johnson2006semantic} have proposed data-driven systems for synthesizing images given  text descriptions and spatial controls based on image matching and blending. Later, neural networks such as conditional GANs have been adopted to directly synthesize images~\cite{isola2017image,chen2017photographic,wang2018high,park2019semantic,sushko2020you,zhu2020sean,huang2022multimodal} but only work on limited domains such as natural landscapes or streets. 
More recently, several works propose to add spatial modulations to a pre-trained text-to-image diffusion model~\cite{ldm, spatext, sketchguide, wang2022pretraining,controlnet,gligen}. For example, SpaText~\cite{spatext} fine-tunes a pre-trained model using a paired dataset of images and segmentation maps. 
ControlNet~\cite{controlnet} and GLIGEN~\cite{gligen} develop more effective fine-tuning methods with adapters.  
Meanwhile, Universal-Guided-Diffusion~\cite{universaldiffusion} and Attention Refocusing~\cite{attentionrefocusing} encourage intermediate outputs to align with the layout condition using guidance functions.
Although these models offer control over image layout, they often require retraining each time for a new type of control or increased inference time.

Several concurrent works also propose training-free spatial control for diffusion models~\cite{he2023localized,phung2023grounded,chen2023training}. We encourage  the readers to check out their works for more details. 
\section{Method}
\label{sec:method}

\begin{figure}[t!]
    \centering
    \begin{tabular}{c}
    \includegraphics[width=0.97\linewidth]{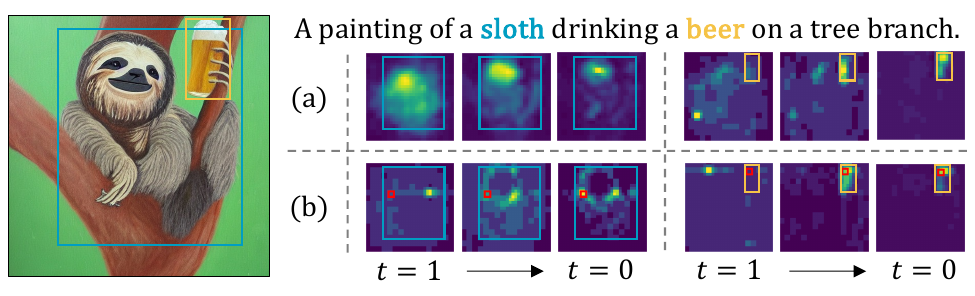} \\
    \end{tabular}
  \caption{
  Visualization of 16 $\times$ 16 attention maps attained from (a) cross-attention and (b) self-attention layers of Stable Diffusion~\cite{ldm}. In (a),  we visualize the cross-attention maps for ``sloth'' and ``beer''.  The objects of interest are outlined with blue and yellow bounding boxes. 
  In (b), we show the attention maps for token keys marked in the red boxes in self-attention layers.
  As the timestep $t$ approaches zero, tokens affiliated with the same object communicate more closely, influencing the image layout.
  }
  \label{fig:concept}
\end{figure}

\begin{figure*}[t!]
    \centering
    \begin{tabular}{c}
    \includegraphics[width=0.97\linewidth]{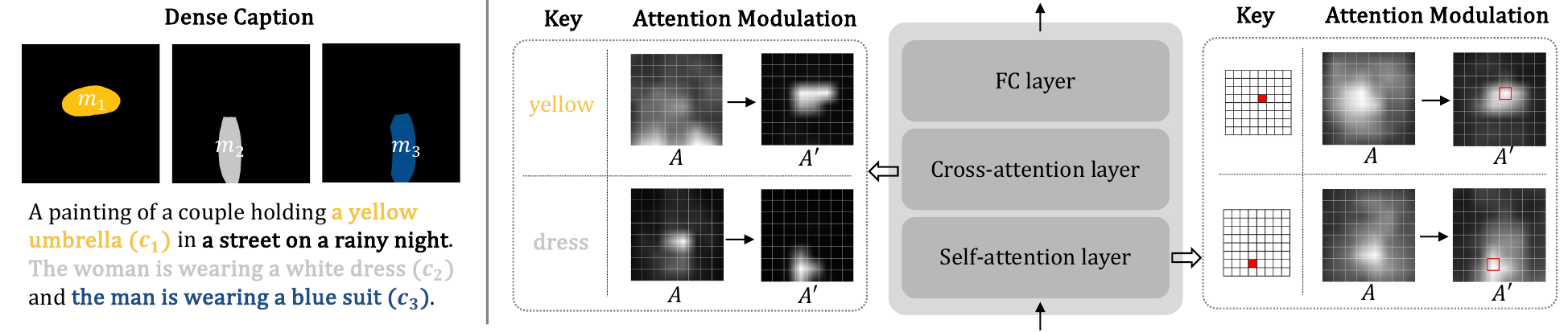} \\
    \end{tabular}
  \caption{
  \textbf{Our attention modulation process.} 
  Our goal is to congregate specific textual features, denoted as $c_n$, into the regions defined by its corresponding layout condition $m_n$.
  At cross-attention layers, we modulate the attention scores between paired image and text tokens in each segment $(c_{n},m_{n})$ to have higher values.
  At self-attention layers, the modulation is applied so that pairs of image tokens belonging to the same object exhibit higher values.
  In attention maps, brighter colors represent higher attention scores. 
  }
  \label{fig:method}
\end{figure*}

Our goal is to improve the text-to-image model's ability to reflect textual and spatial conditions without fine-tuning.
We formally define our condition as a set of $N$ segments ${\{(c_{n},m_{n})\}}^{N}_{n=1}$, where each segment $(c_n,m_n)$ describes a single region, as shown in Figure~\ref{fig:method}.
Here $c_n$ is a non-overlapping part of the full-text caption $c$, and $m_n$ denotes a binary map representing each region. Given the input conditions, we modulate attention maps of all attention layers on the fly so that the object described by $c_n$ can be generated in the corresponding region $m_n$.
To maintain the pre-trained model's generation capacity, we design the modulation to consider original value range and each segment's area.

\subsection{Preliminaries}

\myparagraph{Diffusion Models}learn to generate an image by taking successive denoising steps starting from a random noise map $z_T$~\cite{ddpm,song2020score,sohl2015deep}.
A predefined number of denoising steps determines the degree of noise at each step, and a timestep-dependent noise prediction network ${\epsilon}_\theta$ is trained to predict the noise $\tilde{\epsilon}_{t}$ added to a given input $z_t$. 
Although the earlier models, such as DDPM~\cite{ddpm}, are computationally-expensive, the non-Markovian diffusion method, DDIM~\cite{ddim}, has improved the inference speed by drastically reducing the number of denoising steps.
More recently, latent diffusion models enable even faster generation with improved image quality~\cite{ldm}.
If the task is to generate an image using conditional information $c$, such as a text caption or semantic segmentation map, the condition can be provided via cross-attention layers.
In this work, we experiment with Stable Diffusion~\cite{ldm} and use 50 DDIM denoising steps. %

\begin{table}[t!]
    \centering
    \setlength{\tabcolsep}{10pt}
    \begin{tabular}{lcc}
    &
    \multicolumn{1}{c}{\scriptsize Matched Key} &
    \multicolumn{1}{c}{\scriptsize Unmatched Key}\\
    \hhline{===}
    \multicolumn{1}{c}{\scriptsize {\quad Self-attention Layers (Mean) \quad}} &
    \multicolumn{1}{c}{\scriptsize {0.0148}}&
    \multicolumn{1}{c}{\scriptsize {0.0024}}\\
    \cline{1-3}
    \multicolumn{1}{c}{\scriptsize { Cross-attention Layers (Mean) }} &
    \multicolumn{1}{c}{\scriptsize {0.0410}}&
    \multicolumn{1}{c}{\scriptsize {0.0122}}\\
    \cline{1-3}
    \multicolumn{1}{c}{\scriptsize { Cross-attention Layers (Max) }} &
    \multicolumn{1}{c}{\scriptsize {0.0691}}&
    \multicolumn{1}{c}{\scriptsize {0.0372}}\\
    \cline{1-3}
    \end{tabular}

\caption{
Analysis of attention scores for matched and unmatched keys. First, the  object bounding boxes are detected by YOLOv7~\cite{yolov7}. 
In the context of cross-attention layers, we define a matched key if the key's text token matches the class label of the box. %
In self-attention layers, an image token within the box qualifies as a matched key.
In both layers, matched keys consistently have higher mean and max attention values than unmatched keys. 
}
\label{table:attn_map_value}
\end{table}

\myparagraph{Attention Layers.} 
The attention layer~\cite{bahdanau2014neural,attention} is one of the building blocks of Stable Diffusion~\cite{ldm} that updates intermediate features referencing the context features based on the attention maps $A \in \mathbb{R}^{|\text{queries}| \times |\text{keys}|}$ defined as below,
\begin{equation}
\centering
\begin{gathered}
    A = \mbox{softmax}(\frac{QK^\top}{\sqrt{d}}), 
\end{gathered}
\label{eq:reverse_process}
\end{equation}
in which $Q$ and $K$ are query and key values, each mapped from intermediate features and context features. Here $d$ denotes the length of the key and query features.

In a self-attention layer, intermediate features also serve as context features, allowing us to synthesize globally coherent structures by connecting image tokens across different regions. 
Meanwhile, a cross-attention layer updates conditional on textual features, %
which are encoded from the input text caption $c$ with a CLIP text encoder~\cite{clip}.

\myparagraph{Attention Score Analysis.}

Previous studies show that text-conditioned diffusion models tend to form the position and appearance of each object in early stages and refine details such as color or texture variations in relatively later stages~\cite{percep_prior, ediffi, ernie, prompt_to_prompt}.
We show a similar trend by analyzing 16 $\times$ 16 attention maps produced from Stable Diffusion~\cite{ldm}.
We first synthesize images using 250 captions from the MS-COCO validation set~\cite{mscoco}. As shown in Figure~\ref{fig:concept}, 
attention maps tend to resemble the image layout as the generation proceeds.
Table~\ref{table:attn_map_value} shows a quantitative analysis of attention scores, revealing distinct differences based on whether query-key pairs belong to the same object for both cross-attention and self-attention layers. %
This result suggests that query-key pairs belonging to the same object tend to have larger scores  during the generation process. 

\subsection{Layout-guided Attention Modulation}

The analysis results on attention maps motivate us to intervene in the generation process and modulate original scores to better reflect textual and layout conditions. %
Specifically, we modulate the attention maps as below,
\begin{equation}
\centering
\begin{gathered}
    A' = \mbox{softmax}(\frac{QK^\top + M}{\sqrt{d}}), \\
    M = \lambda_{t} \cdot R \odot M_{\text{pos}} \odot (1-S)\\
    \qquad - \lambda_t \cdot (1-R) \odot M_{\text{neg}} \odot (1-S), 
\end{gathered}
\label{eq:am}
\end{equation}
where the query-key pair condition map $R \in \mathbb{R}^{|\text{queries}| \times |\text{keys}|}$ defines whether to increase or decrease the attention score for a particular pair.
If two tokens belong to the same segment, they form a positive pair, and their attention score will be increased. If not, they form a negative pair, with their attention decreased.

Due to the difference between cross-attention and self-attention layers, $R$ is defined separately for the two layers.
We introduce the matrices $ M_{\text{pos}}, M_{\text{neg}} \in \mathbb{R}^{|\text{queries}| \times |\text{keys}|}$  to consider the original value range, which aims to preserve the pre-trained model's generation capacity. To further adjust the degree of modulation according to the size of each object, we calculate the matrix $S \in \mathbb{R}^{|\text{queries}| \times |\text{keys}|}$ that indicates the segment area for each image query token. %

Finally, we observe that a large modification may deteriorate the image quality as the timestep $t$ approaches zero. Therefore, we use a scalar $\lambda_t$ to adjust the degree of modulation using the power function as below, \begin{equation}
\centering
\begin{gathered}
    \lambda_t := w \cdot t^p, 
\end{gathered}
\label{eq:lambda}
\end{equation}
where the timestep $t \in  [0, 1]$ has been normalized. We use different co-efficients $w^c$ and $w^s$ for cross-attention and self-attention layers.

\myparagraph{Attention Modulation at Cross-attention Layers.}

In cross-attention layers, intermediate image features are updated according to textual features, which construct objects' appearance and layouts.
The degree and location to which they are reflected are determined by the attention scores between image tokens and text tokens.
Hence, we modify cross-attention maps to induce certain textual features congregate in a specific region according to its corresponding layout condition $m_n$, with the query-key pair condition map $R^{\text{cross}}$ defined as below,
\begin{equation}
\centering
\begin{gathered}
    R_{:j}^{\text{cross}} := \Bigl\{
        \begin{array}{ll}
        \vec{0}, & \;\; \text{if } \qquad \vec{k}[j] = 0 \\
        \vec{m}_{\vec{k}[j]}, & \;\; \text{otherwise}\\
        \end{array}, \\
\end{gathered}
\label{eq:am_cross}
\end{equation}
where $\vec{m}_n$ is a binary map flattened into a vector and resized to match the spatial resolution of each layer. Here $\vec{k} \in \mathbb{R}^{|\text{keys}|}$ is a vector that maps a token index to the segment index. %
If the $j$-th text token is not part of any of the $N$ segments, $\vec{k}[j]$ is set to zero.

\myparagraph{Attention Modulation at Self-attention Layers.}
Self-attention layers allow intermediate features to interact with each other to create a globally coherent result.
Our attention modulation is designed to restrict communication between tokens of different segments, thereby preventing the mixing of features of distinct objects.
Specifically, we increase the attention scores for tokens in the same segment and decrease it for those in different segments. %
The query-key pair condition map $R^{\text{self}}$ for this objective is defined as below,
\begin{equation}
\centering
\begin{gathered}
    R_{ij}^{\text{self}} := \Bigl\{
        \begin{array}{ll}
        1, & \;\; \text{if } \;\; \exists \ n \;\; \text{s.t.} \;\; \vec{m}_n[i] = 1 \ \text{and} \ \vec{m}_n[j] = 1\\
        0, & \;\; \text{otherwise}
        \end{array}. \\
\end{gathered}
\label{eq:am_self}
\end{equation}

\myparagraph{Value-range Adaptive Attention Modulation.}

Since our method alters the original denoising process,  it may potentially compromise the image quality of the pre-trained model.
To mitigate this risk, we modulate values according to the range of original attention scores.
We calculate the following matrices that identify each query's maximum and minimum values, ensuring the modulated values stay close to the original range.
Therefore, the adjustment is proportional to the difference between the original values and either the maximum value (for positive pairs) or the minimum value (for negative pairs).
\begin{equation}
\centering
\begin{gathered}
    M_{\text{pos}} = \mbox{max}(QK^\top)-QK^\top, \\
    M_{\text{neg}} = QK^\top - \mbox{min}(QK^\top). \\
\end{gathered}
\label{eq:am_range}
\end{equation}
The $\mbox{max}$ and $\mbox{min}$ operations return the maximum and minimum values for each query, initially producing a  vector with dimensions $\mathbb{R}^{|\text{queries}| \times 1}$.
To restore the original dimensions $\mathbb{R}^{|\text{queries}| \times |\text{keys}|}$, we replicate values along the key-axis.

\myparagraph{Mask-area Adaptive Attention Modulation.}

We observe a noticeable quality degradation when there is a large area difference between segments.
Specifically, if one segment's area is much smaller than others, our method may fail to generate realistic images. 
To resolve this, we use the matrix $S$ in Equation~\ref{eq:am} to automatically adjust the modulation degree according to the area of each segment: increase the degree for small segments and decrease for large segments. To calculate the matrix $S$, we first compute the area percentage of the mask that each query token belongs to, and then replicate the values along the key axis.

\myparagraph{Implementation Details.}

We use Stable Diffusion~\cite{ldm}, trained on the LAION dataset~\cite{laion}.
In our experiments, we only apply the attention modulation to the initial denoising steps ($t=1 \sim 0.7$), as we observe no clear improvement beyond that.
The hyperparameters in Equation~\ref{eq:lambda} are set as follows:  $p=5$, $w^c=1$  for cross-attention layers, and $w^s=0.3$ for self-attention layers.
To further increase the effectiveness of our method, we replace parts of textual features with separately encoded ones for each text segment $c_n$.
This strategy is particularly useful when a text caption contains multiple closely related objects, such as a microwave and an oven.

\section{Experiments}
\label{sec:experimets}

\begin{figure*}[t]
    \centering
    \begin{tabular}{cccccc}
     &
     \multicolumn{1}{c}{\makecell{\qquad \qquad \quad \; \small{Dense} \\ \qquad \qquad \quad \; \small{Diffusion (Ours)}}}&
    \multicolumn{1}{c}{\quad \:\: SD-Pww~\cite{ediffi}} &
    \multicolumn{1}{c}{\makecell{\qquad \:\: \small{Composable} \\ \:\: \qquad \small{Diffusion~\cite{compdiff}}}} &
    \multicolumn{1}{c}{\makecell{\qquad \ \small{Structure} \\ \qquad \small{Diffusion~\cite{strucdiff}}}} &
    \multicolumn{1}{c}{\makecell{\quad \small{Stable} \\ \quad \small{Diffusion~\cite{ldm}}}} \\
    
    \multicolumn{6}{c}{\includegraphics[width=0.97\linewidth]{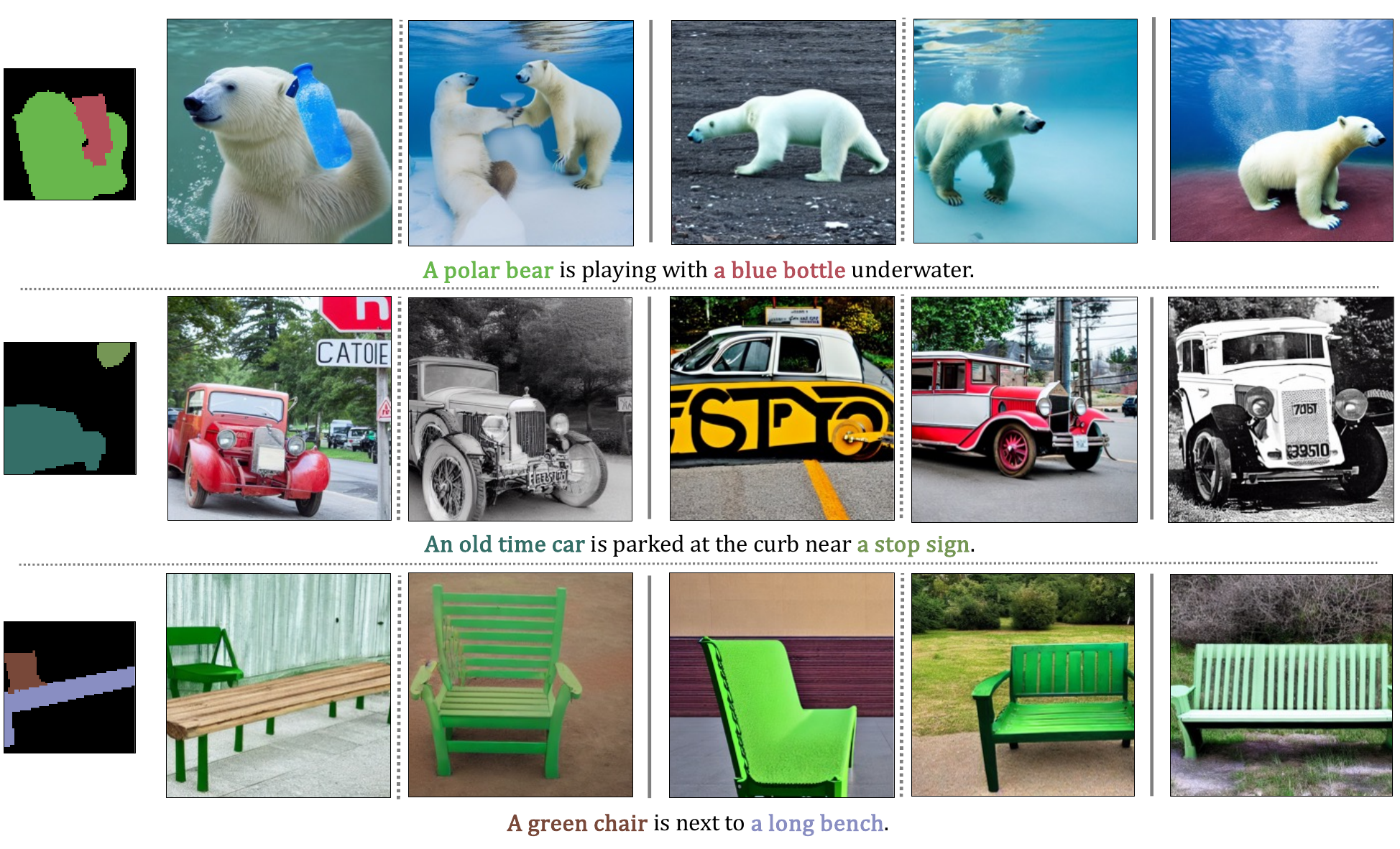}} \\
    
    \end{tabular}
  \vspace*{-0.5 cm}
  \captionof{figure}{
  \textbf{Comparisons with other training-free methods based on Stable Diffusion~\cite{ldm}.} 
  Each image is generated with the same dense caption for all methods.
  However, only SD-Pww~\cite{ediffi} and our method \ours support segmentation maps for layout control. Our method aligns more closely with the input mask than SD-Pww~\cite{ediffi}. 
  }
  \label{fig:quali_dense}
\end{figure*}

\subsection{Evaluation Setting}

\myparagraph{Baselines.}

We compare our method with various \textit{training-free} methods, which aim to improve the fidelity of the pre-trained Stable Diffusion~\cite{ldm} when using dense captions.
\textbf{Composable Diffusion}~\cite{compdiff} predicts noises for each phrase separately and merges them at every timestep, leading to a longer inference time proportional to the number of phrases. 
\textbf{Structure Diffusion}~\cite{strucdiff} employs a language parser to enhance the value features of all parsed phrases at every cross-attention layer. %
eDiffi-Pww~\cite{ediffi} manipulates the attention scores for specific query-key pairs according to an additional layout condition.
However, since the eDiffi model is not publicly available, we conduct experiments with the Pww implemented on Stable Diffusion (\textbf{SD-Pww}).

\begin{figure*}[t]
    \centering
    \begin{tabular}{cccccc}

    &
    \multicolumn{1}{c}{\makecell{\qquad \qquad \quad \; \small{Dense} \\ \qquad \qquad \quad \; \small{Diffusion (Ours)}}}&
    \multicolumn{1}{c}{\quad \:\: SD-Pww~\cite{ediffi}} &
    \multicolumn{1}{c}{\qquad \quad  MAS~\cite{makeascene}} &
    \multicolumn{1}{c}{\qquad \; SpaText~\cite{spatext}} &
    \multicolumn{1}{c}{\makecell{\quad \ \small{Stable} \\ \quad \ \small{Diffusion~\cite{ldm}}}} \\
    \multicolumn{6}{c}{\includegraphics[width=0.97\linewidth]{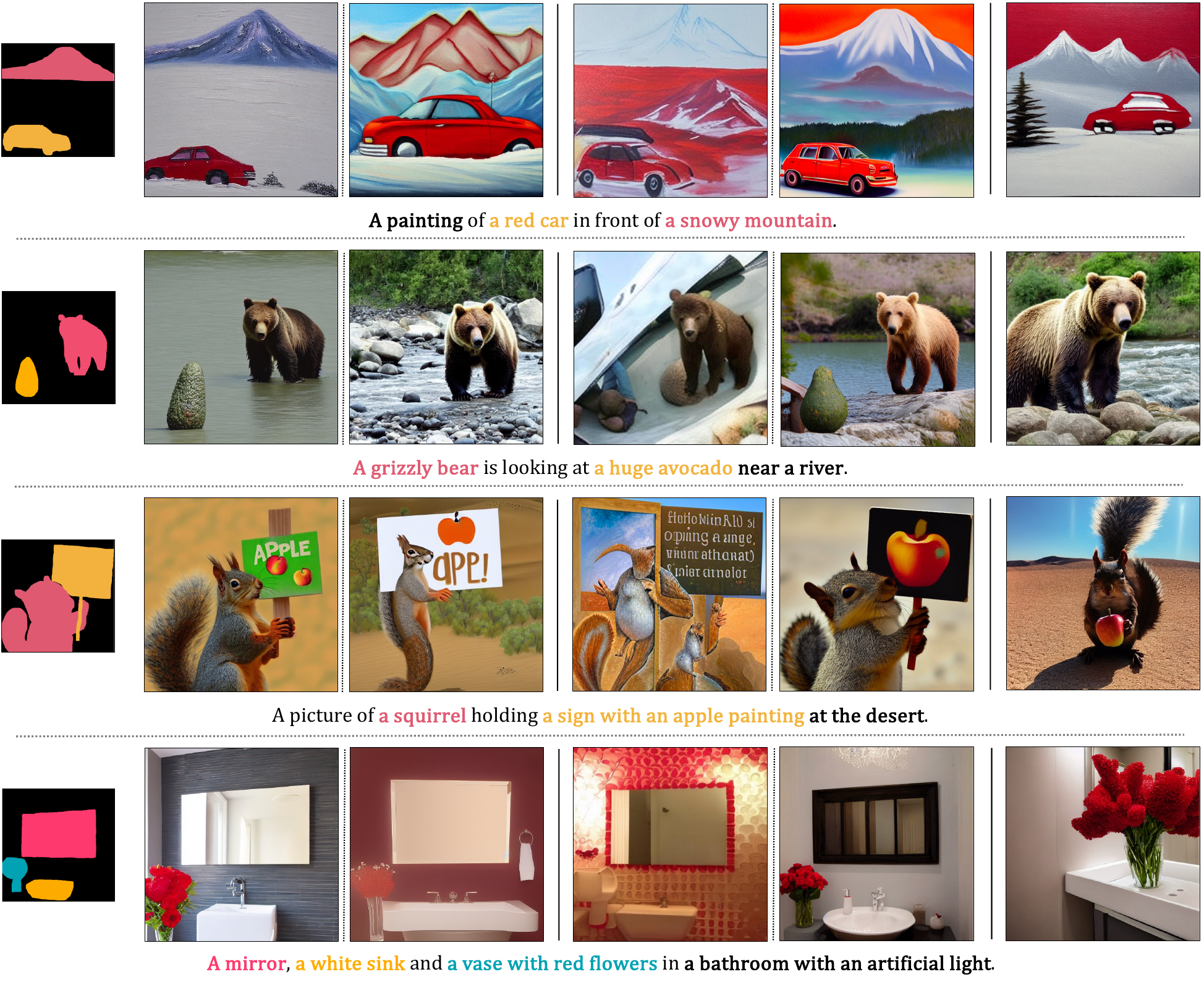}} \\
    \end{tabular}
  \caption{
  \textbf{Comparison with different layout-guided text-to-image methods.}
  While MAS~\cite{makeascene} and SpaText~\cite{spatext} are specifically trained for layout control, SD-Pww~\cite{ediffi} and \ours (ours) are training-free methods based on pre-trained Stable Diffusion models~\cite{ldm}. 
   Nevertheless, our results adhere to layout conditions comparably to SpaText~\cite{spatext}, and even outperform MAS~\cite{makeascene} in many cases. 
  }
  \label{fig:quali_layout}
\end{figure*}

\myparagraph{Metrics.} 
We evaluate each method regarding two criteria:  fidelity to the text prompt and alignment with the layout condition. 
For text prompt, we calculate the CLIP-Score~\cite{clip-score}, which measures the distance between input textual features and generated image features, and the SOA-I score~\cite{soa} that uses YOLOv7~\cite{yolov7} to check if the described objects are present in the generated image.
 In terms of layout alignment, we compare IoU scores of a segmentation map predicted by YOLOv7~\cite{yolov7}, with respect to the given layout condition.
We further evaluate CLIP-scores on cropped object images (Local CLIP-score~\cite{spatext}) to check whether generated objects followed detailed descriptions.
As Composable Diffusion~\cite{compdiff} and Structure Diffusion~\cite{strucdiff} do not take layout conditions, they are excluded for a fair comparison.

\myparagraph{Datasets.}

We curate a new evaluation dataset containing detailed descriptions for each segment. 
Specifically, we select 250 samples with two or more unique objects from the MS-COCO  validation set~\cite{mscoco}.
Then we manually replace the class label for each segmentation map with a phrase extracted from the caption; for example, ``dog'' to ``a black and white dog''.
We generate four random images per caption, leading to 1,000 images for each baseline used in the evaluation.

\begin{figure*}[t]
    \centering
    \begin{tabular}{M{16cm}}
    \multicolumn{1}{c}{\includegraphics[width=0.97\linewidth]{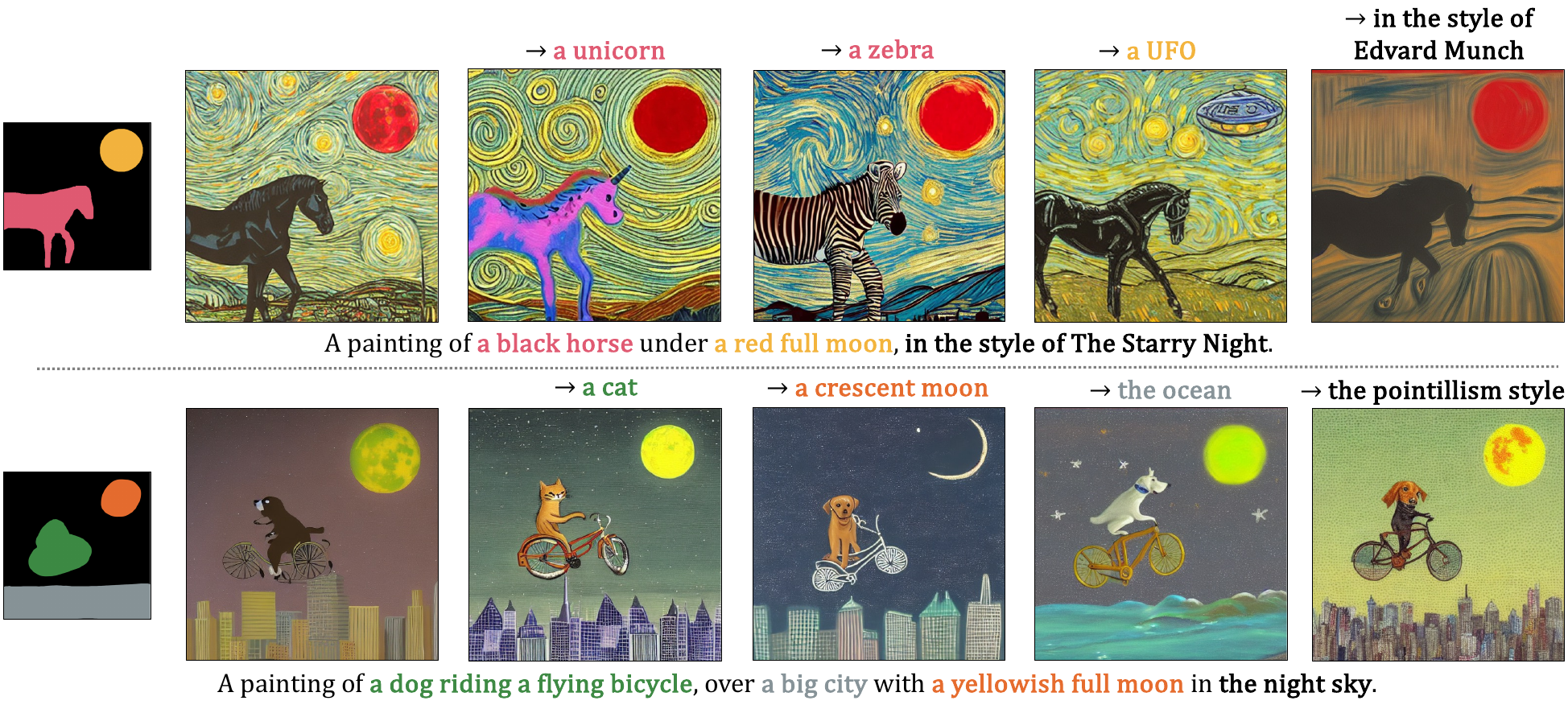}} \\
    \end{tabular}
  \vspace*{-0.2 cm}
  \caption{
We generate images using the same layout condition but with various text prompts, by modifying part of the given textual condition. Our results faithfully adhere to both textual and layout conditions.
  }
  \label{fig:quali_edit}
\end{figure*}

\begin{table}[t]
    \centering
    \setlength{\tabcolsep}{5pt}
    \begin{tabular}{lccc}

    &
    \multicolumn{1}{c}{\makecell{\scriptsize{SOA-I $\uparrow$}}}&
    \multicolumn{1}{c}{\scriptsize{CLIP-Score $\uparrow$}}&
    \multicolumn{1}{c}{\makecell{\scriptsize{Human} \\ \scriptsize{Preference $\uparrow$}}}\\
    
    \hhline{====}

    \multicolumn{1}{c}{\scriptsize {Real Images}}&
    \multicolumn{1}{c}{\scriptsize {93.07}}&
    \multicolumn{1}{c}{\scriptsize {0.2821}}&
    \multicolumn{1}{c}{\scriptsize {-}}\\

    \cline{1-4}

    \multicolumn{1}{c}{\scriptsize {Stable Diffusion~\cite{ldm}}} &
    \multicolumn{1}{c}{\scriptsize {73.08 $\pm$ 1.54}}&
    \multicolumn{1}{c}{\scriptsize {0.2732 $\pm$ 0.0016}}&
    \multicolumn{1}{c}{\scriptsize {$62\%$}}\\

    \multicolumn{1}{c}{\scriptsize {Composable Diffusion~\cite{compdiff}}}&
    \multicolumn{1}{c}{\scriptsize {55.88 $\pm$ 0.78}}&
    \multicolumn{1}{c}{\scriptsize {0.2507 $\pm$ 0.0023}}&
    \multicolumn{1}{c}{\scriptsize {$83\%$}}\\

    \multicolumn{1}{c}{\scriptsize {Structure Diffusion~\cite{strucdiff}}}&
    \multicolumn{1}{c}{\scriptsize {70.97 $\pm$ 1.08}}&
    \multicolumn{1}{c}{\scriptsize {0.2745 $\pm$ 0.0011}}&
    \multicolumn{1}{c}{\scriptsize {$58\%$}}\\

    \cline{1-4}

    \multicolumn{1}{c}{\scriptsize {SD-Pww~\cite{ediffi}}} &
    \multicolumn{1}{c}{\scriptsize {73.92 $\pm$ 1.84}}&
    \multicolumn{1}{c}{\scriptsize {0.2800 $\pm$ 0.0005}}&
    \multicolumn{1}{c}{\scriptsize {$53\%$}}\\

    \multicolumn{1}{c}{\scriptsize {\ours (Ours)}}&
    \multicolumn{1}{c}{\scriptsize {\bf{77.61 $\pm$ 1.75}}}&
    \multicolumn{1}{c}{\scriptsize {\bf{0.2814 $\pm$ 0.0005}}}&
    \multicolumn{1}{c}{\scriptsize {-}}\\
    
    \cline{1-4}
    \end{tabular}   
  \caption{
  Quantitative evaluation results on the fidelity to textual condition.
  Our method achieves the best performance regarding both automatic metrics and user study. The human preference percentage shows how often AMT participants prefer our results over the baselines. 
  }
  \label{table:quant_textual}

    \centering
    \setlength{\tabcolsep}{6pt}
    \begin{tabular}{lccc}
    
    &
    \multicolumn{1}{c}{\scriptsize{IoU $\uparrow$}} &
    \multicolumn{1}{c}{\makecell{\scriptsize{Local} \\ \scriptsize{CLIP-Score $\uparrow$}}} &
    \multicolumn{1}{c}{\makecell{\scriptsize{Human} \\ \scriptsize{Preference $\uparrow$}}} \\
    
    \hhline{====}

    \multicolumn{1}{c}{\scriptsize {Real Images}} &
    \multicolumn{1}{c}{\scriptsize {59.13}}&
    \multicolumn{1}{c}{\scriptsize {0.2163}}&
    \multicolumn{1}{c}{\scriptsize {-}}\\

    \cline{1-4}

    \multicolumn{1}{c}{\scriptsize {SD-Pww~\cite{ediffi}}} &
    \multicolumn{1}{c}{\scriptsize {23.76 $\pm$ 0.50}}&
    \multicolumn{1}{c}{\scriptsize {0.2125 $\pm$ 0.0003}}&
    \multicolumn{1}{c}{\scriptsize { $63\%$ }}\\

    \multicolumn{1}{c}{\scriptsize {\ours (Ours)}} &
    \multicolumn{1}{c}{\scriptsize \bf{{34.99 $\pm$ 1.13}}}&
    \multicolumn{1}{c}{\scriptsize {\bf{0.2176 $\pm$ 0.0007}}}&
    \multicolumn{1}{c}{\scriptsize {-}}\\
    
    \cline{1-4}
    \end{tabular}   
  \caption{
  Quantitative evaluation results on the fidelity to layout condition.
  We compare only with SD-Pww~\cite{ediffi} since it is the only baseline that uses segmentation maps. $63\%$ of AMT participants prefer our results over those from SD-Pww. 
  }
  \label{table:quant_layout}
\end{table}

\myparagraph{User study.}

We conduct a user study using Amazon Mechanical Turk. 
For each task, we present users with two sets of 4 images along with the same input conditions. 
They are asked to choose the better set based on either of the following criteria: fidelity to the textual condition while reflecting detailed descriptions of key objects or fidelity to the layout condition with accurate depiction of objects. 
We present each pair in a random order and collect three ratings from unique users.
The scores in Table~\ref{table:quant_textual} and~\ref{table:quant_layout} show the percentage of users who prefer \ours over a baseline. 50\% means that \ours and a baseline have the same preference, and values more than 50\% indicate that more users select \ours instead of the  baseline.

\subsection{Results}

\myparagraph{Evaluation on the fidelity to textual condition.}
In Figure~\ref{fig:quali_dense}, we compare \ours with all baselines for images generated with dense captions.
While baseline methods sometimes omit one or more objects described in the text captions, our results are more faithful to both textual and layout conditions.
In particular, the comparison with SD-Pww~\cite{ediffi} highlights the effectiveness of our training-free modulation method.

The quantitative evaluation results in Table~\ref{table:quant_textual} reveal a consistent trend.
\ours outperforms others in terms of both automatic and human evaluation.
However, SOA-I seems to be loosely correlated with the human evaluation results due to the domain gap between LAION~\cite{laion} and MS-COCO~\cite{mscoco}, used for training Stable Diffusion~\cite{ldm} and YOLOv7~\cite{yolov7} respectively. Interestingly, the performance tends to suffer significantly when the inference method is altered too much from the original one, as seen in the case of Composable Diffusion~\cite{compdiff}.

\myparagraph{Evaluation on the fidelity to layout condition.}

To evaluate the fidelity to layout conditions, we only compare with the results of SD-Pww~\cite{ediffi}, the only baseline that can control an image layout.
Figures~\ref{fig:quali_dense},~\ref{fig:quali_layout} and Table~\ref{table:quant_layout} show that \ours outperforms SD-Pww by a large margin.
SD-Pww not only fails to reflect layout conditions faithfully but also tends to mix different object features or omits key objects.
In particular, the substantial difference in IoU scores shows that \ours is more effective in reflecting layout conditions.
Figure~\ref{fig:quali_edit} shows that our method responds well to various conditions created by changing part of a given textual condition, such as object types or image styles while maintaining the original layout condition.

\begin{table}[t]
    \centering
    \setlength{\tabcolsep}{4pt}
    \begin{tabular}{lcccc}
    
    &
    \multicolumn{1}{c}{\scriptsize{SOA-I $\uparrow$}}&
    \multicolumn{1}{c}{\scriptsize{CLIP-Score $\uparrow$}}&
    \multicolumn{1}{c}{\scriptsize{IoU $\uparrow$}}&
    \multicolumn{1}{c}{\makecell{\scriptsize{Local} \\ \scriptsize{CLIP-Score $\uparrow$}}}\\
    
    \hhline{=====}

    \cline{1-5}

    \multicolumn{1}{c}{\scriptsize {w/o (a)}} &
    \multicolumn{1}{c}{\scriptsize {72.27}}&
    \multicolumn{1}{c}{\scriptsize {0.2785}}&
    \multicolumn{1}{c}{\scriptsize {18.35}}&
    \multicolumn{1}{c}{\scriptsize {0.2050}}\\

    \multicolumn{1}{c}{\scriptsize {w/o (b)}} &
    \multicolumn{1}{c}{\scriptsize {75.60}}&
    \multicolumn{1}{c}{\scriptsize {0.2790}}&
    \multicolumn{1}{c}{\scriptsize {29.57}}&
    \multicolumn{1}{c}{\scriptsize {0.2152}}\\

    \multicolumn{1}{c}{\scriptsize {w/o (c)}} &
    \multicolumn{1}{c}{\scriptsize {76.51}}&
    \multicolumn{1}{c}{\scriptsize {0.2801}}&
    \multicolumn{1}{c}{\scriptsize {29.88}}&
    \multicolumn{1}{c}{\scriptsize {0.2136}}\\

    \multicolumn{1}{c}{\scriptsize {w/o (d)}} &
    \multicolumn{1}{c}{\scriptsize {73.62}}&
    \multicolumn{1}{c}{\scriptsize {0.2808}}&
    \multicolumn{1}{c}{\scriptsize {\bf{38.00}}}&
    \multicolumn{1}{c}{\scriptsize {\bf{0.2191}}}\\

    \cline{1-5}

    \multicolumn{1}{c}{\scriptsize {\ours}} &
    \multicolumn{1}{c}{\scriptsize {\bf{77.61}}}&
    \multicolumn{1}{c}{\scriptsize {\bf{0.2814}}}&
    \multicolumn{1}{c}{\scriptsize {34.99}}&
    \multicolumn{1}{c}{\scriptsize {0.2176}}\\  
    
    \cline{1-5}
    
    \end{tabular}   
    \vspace{-5pt}
  \caption{\textbf{Quantitative evaluations on each component. }
  We evaluate several variants, in which the following components are removed from our full method: (a) Attention Modulation at Cross-attention Layers, 
(b) Attention Modulation at Self-attention Layers,
(c) Value-range Adaptive Attention Modulation, 
and (d) Mask-area Adaptive Attention Modulation.
The results suggest that each component contributes  to the final performance of \ours.
  }
  \label{table:quant_abla}
  \vspace{-5pt}
\end{table}

\begin{figure*}[t!]
    \centering
    \centering
    \begin{tabular}{ccccc}

    \multicolumn{1}{c}{\qquad\qquad\qquad\;\ours}&
    \multicolumn{1}{c}{\qquad\quad w/o (a)} &
    \multicolumn{1}{c}{\qquad\qquad\;\: w/o (b)} &
    \multicolumn{1}{c}{\qquad\qquad\;\: w/o (c)} &
    \multicolumn{1}{c}{\qquad w/o (d)} \\
    
    \multicolumn{5}{c}{\includegraphics[width=0.97\linewidth]{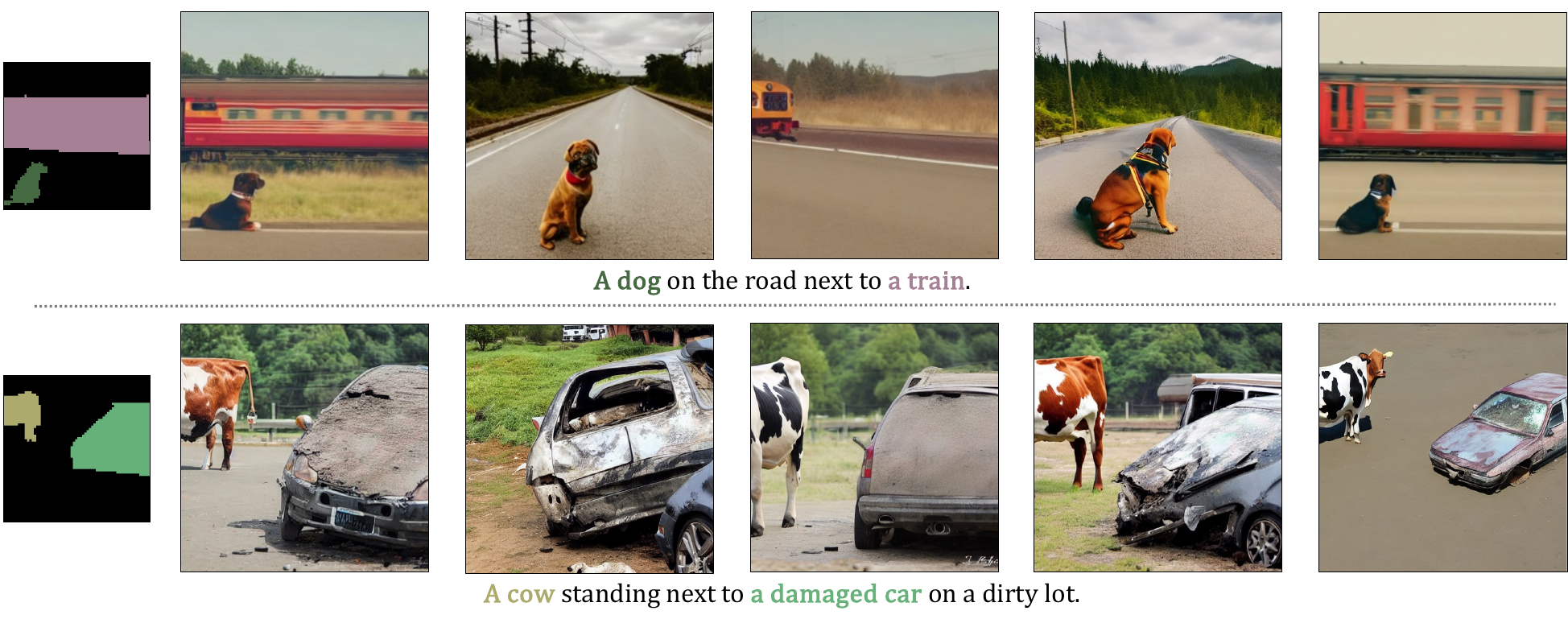}} \\
    \end{tabular}
  \vspace{-10pt}
  \caption{
  \textbf{Ablation study.} 
  We show some example results when various components are ablated from our full method. We define the following components as: 
  (a) Attention Modulation at Cross-attention Layers, (b) Attention Modulation at Self-attention Layers, (c) Value-range Adaptive Attention Modulation, and (d) Mask-area Adaptive Attention Modulation. 
  All images are generated with the same initial noise map.
   We can infer that all components contribute to improving the fidelity of Stable Diffusion to the given conditions.
  }
  \label{fig:abla}
\vspace{-7pt}
\end{figure*}

\begin{figure}
\centering

    \captionsetup[subfigure]{labelformat=empty}
    \begin{subfigure}
    \centering
    \includegraphics[width=1.0\linewidth]{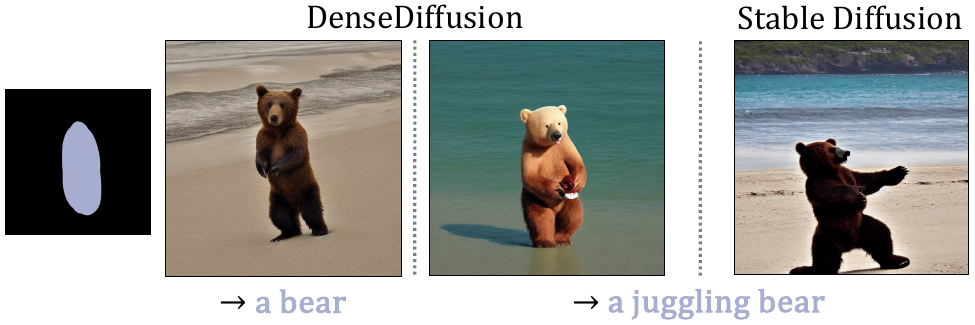}
      \vspace{-15pt}
    \caption*{(a) Incorrect content generation }
    \end{subfigure}%
    \begin{subfigure}
    \centering
    \includegraphics[width=1.0\linewidth]{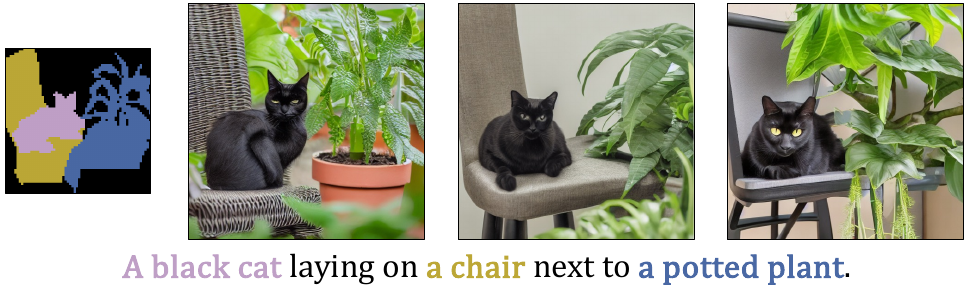}
    \vspace{-15pt}
    \caption*{(b) Fine-grained user masks}
    \end{subfigure}%
\vspace{-5pt}
\caption{\textbf{Limitations and failure cases.} (a) Given captions, ``a photo of (a bear / a juggling bear) at the beach'', our method fails in the latter case, since the Stable Diffusion itself cannot generate an image with the correct appearance.  (b) As both self-attention and cross-attention layers are fairly coarse, our method fails to follow fine details of the segment condition, such as the shape of leaves.}
\label{fig:limitation}
\end{figure}

\myparagraph{Comparison to layout-conditioned models.}

To highlight that \ours is effective even as a training-free method, we further compare with MAS~\cite{makeascene} and SpaText~\cite{spatext}, both of which are text-to-image models trained with layout conditions. %
MAS~\cite{makeascene} uses tokenized semantic segmentation maps as an additional condition, and SpaText~\cite{spatext} fine-tunes Stable Diffusion~\cite{ldm} with spatially splatted CLIP image features according to a layout condition. 
Since these models are not publicly available, we  use the examples presented in the original SpaText paper.
Figure~\ref{fig:quali_layout} shows that \ours can reflect layout conditions comparably and even outperforms MAS~\cite{makeascene} for diverse concepts. 
Please refer to Appendix~\ref{appendix} for more additional comparisons and results.

\myparagraph{Ablation Study.}

Below we evaluate each component used in our \ours:
(a) Attention Modulation at Cross-attention Layers, 
(b) Attention Modulation at Self-attention Layers,
(c) Value-range Adaptive Attention Modulation, 
and (d) Mask-area Adaptive Attention Modulation. 

We first present visual results from our ablation study in Figure~\ref{fig:abla}. 
All images in the same row are generated from the same initial noise map. %
As shown in columns \underline{w/o (a)} and \underline{w/o (b)}, attention modulations in both cross-attention and self-attention layers are crucial for satisfying both textual and layout conditions. The images in column \underline{w/o (c)} show that value-range adaptive modulation improves the fidelity of the method to given conditions further.
Finally, according to column \underline{w/o (d)},  the method adheres to the conditions but produces images with monotonous textures. %

In Table~\ref{table:quant_abla}, we evaluate various ablated methods using automatic metrics. The results show that, except for component (d), the removal of each component leads to a significant drop in all metric scores.
As indicated by the significant drops in IoU scores, the ablated methods fail to achieve high fidelity to layout conditions. %
The largest performance degradation is observed when attention modulation at cross-attention layers is omitted, as textual features play a crucial role in constructing the image layout. Furthermore, modulating scores without considering the value-range also results in drops across all metrics.
These findings collectively confirm that our attention modulation is highly effective in improving the fidelity to given conditions without compromising the generation capability of the pre-trained model.

Regarding component (d),  we interpret this exception as a result of the ablated method's tendency to create a monotonous background, as shown in Figure~\ref{fig:abla}.  While it may seem distant from a realistic image, it is easier for the segmentation model to predict a segmentation map.
Hence, it helps to get good scores on metrics related to layout conditions, but it does not always satisfy the textual conditions. %

\section{Conclusion}
\label{sec:conclusion}

In this work, we have proposed \ours, a training-free method that improves the fidelity of a pre-trained text-to-image model to dense captions, and enable image layout control.
Our finding shows that considering the value range and segment size significantly improves our attention modulation method.  %
Experimental results show that \ours outperform other methods on various evaluation metrics. 
Notably, our training-free method provides comparable layout control to existing models that are specifically trained for this task.

\myparagraph{Limitations.}
\ours has several limitations. First, our method is highly dependent on the capacity of its base model, Stable Diffusion~\cite{ldm}. 
As shown in Figure~\ref{fig:limitation}a, our method fails to produce certain objects, such as a juggling bear, if Stable Diffusion is unable to produce them itself.  

Second, our method struggles with fine-grained input masks with thin structures, as both self-attention and cross-attention layers are fairly coarse. 
As examples shown in Figure~\ref{fig:limitation}b, our method fails to follow fine details of the segment condition, such as the shape of leaves.

\myparagraph{Acknowledgments.} We thank Or Patashnik and Songwei Ge for paper proofreading and helpful comments. 
This work was experimented on the NAVER Smart Machine Learning (NSML) platform~\cite{kim2018nsml}.
{\small
\bibliographystyle{ieee_fullname}
\bibliography{ieee_fullname}
}

\newpage
\clearpage
\appendix

\section{Additional Comparison}
\label{appendix}
Here, we present additional qualitative results generated by concurrent works on several dense captions. %
Specifically, three captions are selected from MS-COCO~\cite{mscoco}, and the other three are curated  by the authors.  %

\subsection{Baselines}

We compare to the following baselines, which also address the limitations of pre-trained text-to-image diffusion models, including (1) object omission or the mixing of visual features from different objects and (2) the lack of spatial controls.  All the methods are implemented on or fine-tuned using the Stable Diffusion~\cite{ldm}.

Both \textbf{Attend-and-Excite}~\cite{attendnexcite} and \textbf{Universal-Guided-Diffusion}~\cite{universaldiffusion} modify the inference algorithm with additional objectives. %
Attend-and-Excite~\cite{attendnexcite} encourages the subject text tokens to be active in cross-attention layers. 
This method does not support additional layout conditions. 
Universal-Guided-Diffusion~\cite{universaldiffusion} utilizes off-the-shelf segmentation network, MobileNetV3~\cite{howard2019searching}, to incorporate additional guidance. 
Specifically, it induces the mask inferred from the predicted clean image to match the given conditional mask. 
However, since MobileNetV3 is trained on MS-COCO dataset, they cannot be applied to our curated prompts that contain out-of-domain objects. We only report their results on our MS-COCO prompts. 
Meanwhile, \textbf{MultiDiffusion}~\cite{multidiffusion} adopts an independent denoising process for each phrase, and use the layout masks to blend the predicted noises. 
These three training-free methods are much slower than the original inference method, as their inference time is proportional to the number of objects.

\subsection{Results}

\begin{figure*}[t]
    \centering
    \begin{tabular}{c}
    \multicolumn{1}{c}{\includegraphics[width=0.97\linewidth]{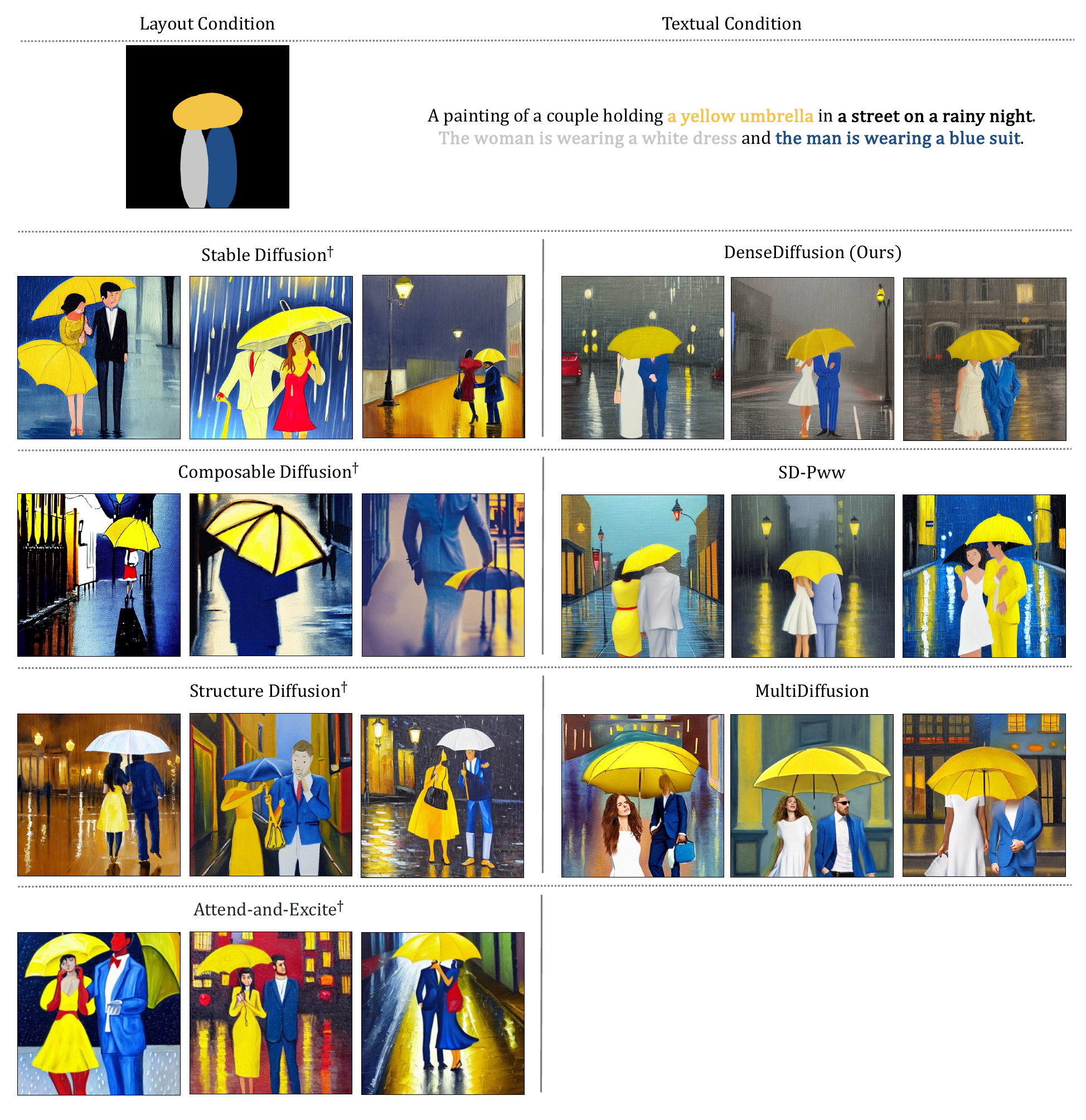}} \\
    \end{tabular}
  \caption{
  Qualitative comparison with additional methods.
  Methods denoted with $^{\dagger}$ receive textual condition only.
  }
  \label{fig:custom_0}
\end{figure*}

\begin{figure*}[t]
    \centering
    \begin{tabular}{c}
    \multicolumn{1}{c}{\includegraphics[width=0.97\linewidth]{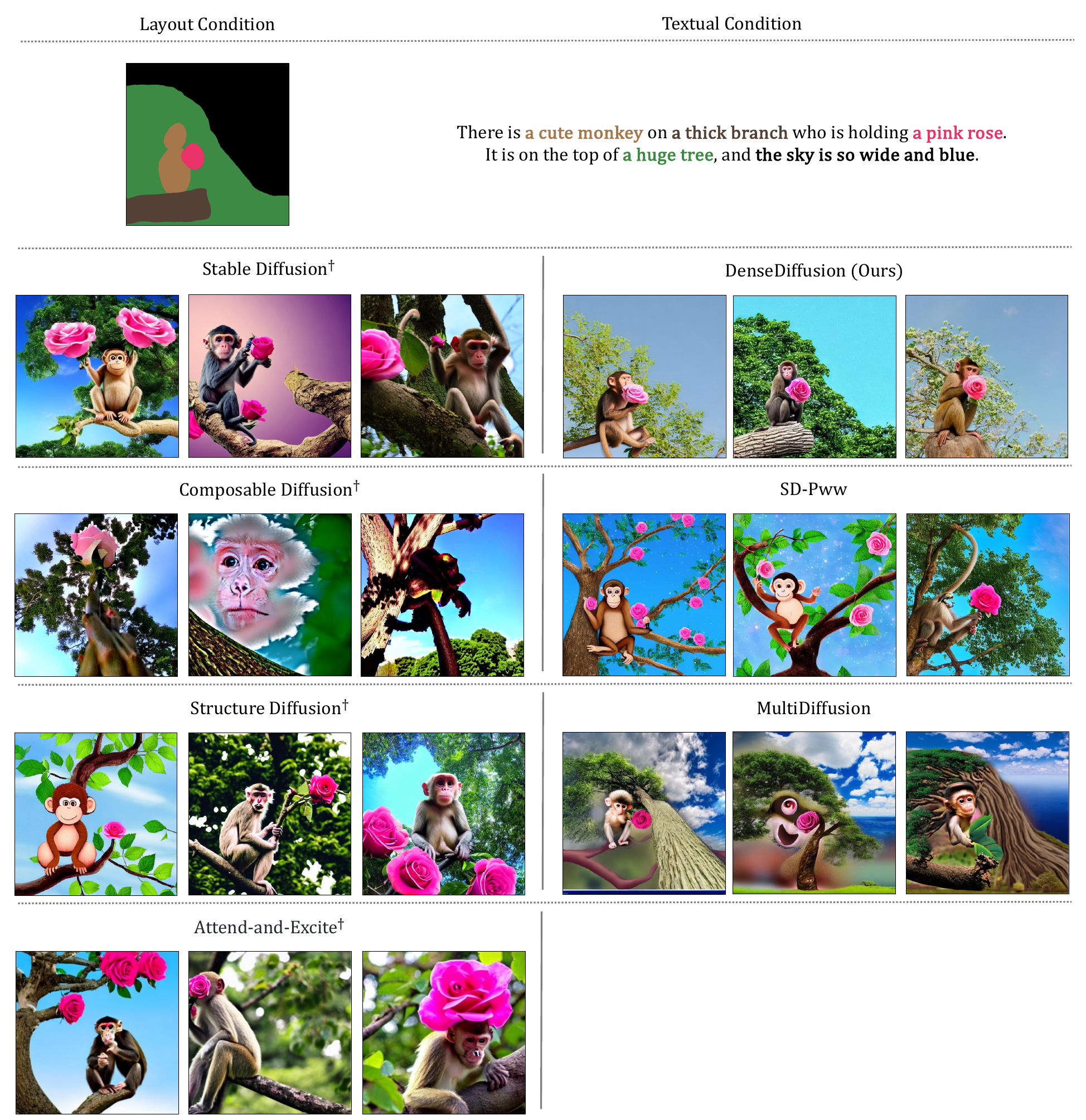}} \\
    \end{tabular}
  \caption{
  Qualitative comparison with additional methods.
  Methods denoted with $^{\dagger}$ receive textual condition only.
  }
  \label{fig:custom_1}
\end{figure*}

\begin{figure*}[t]
    \centering
    \begin{tabular}{c}
    \multicolumn{1}{c}{\includegraphics[width=0.97\linewidth]{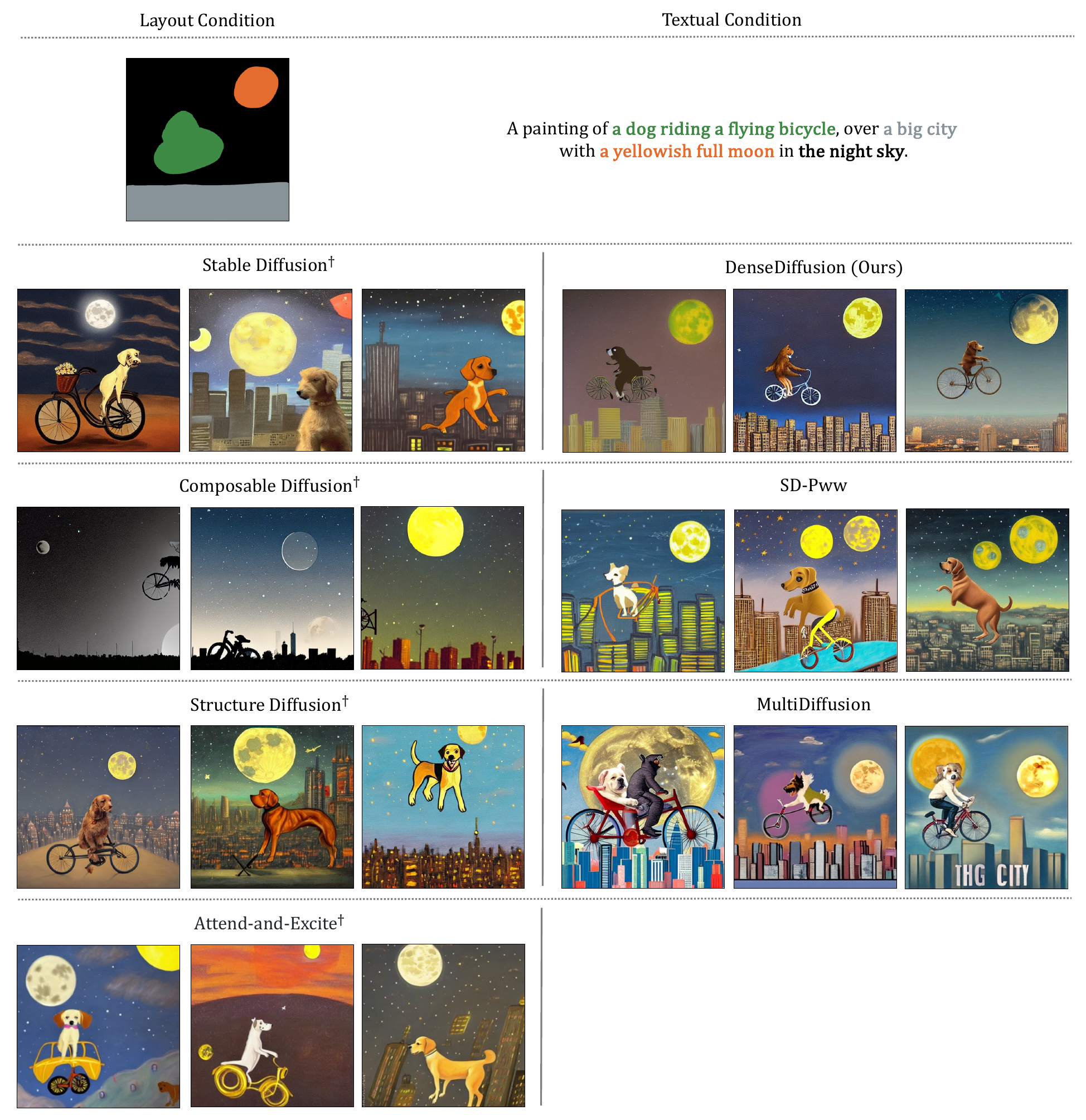}} \\
    \end{tabular}
  \caption{
  Qualitative comparison with additional methods.
  Methods denoted with $^{\dagger}$ receive textual condition only.
  }
  \label{fig:custom_2}
\end{figure*}

\begin{figure*}[t]
    \centering
    \begin{tabular}{c}
    \multicolumn{1}{c}{\includegraphics[width=0.97\linewidth]{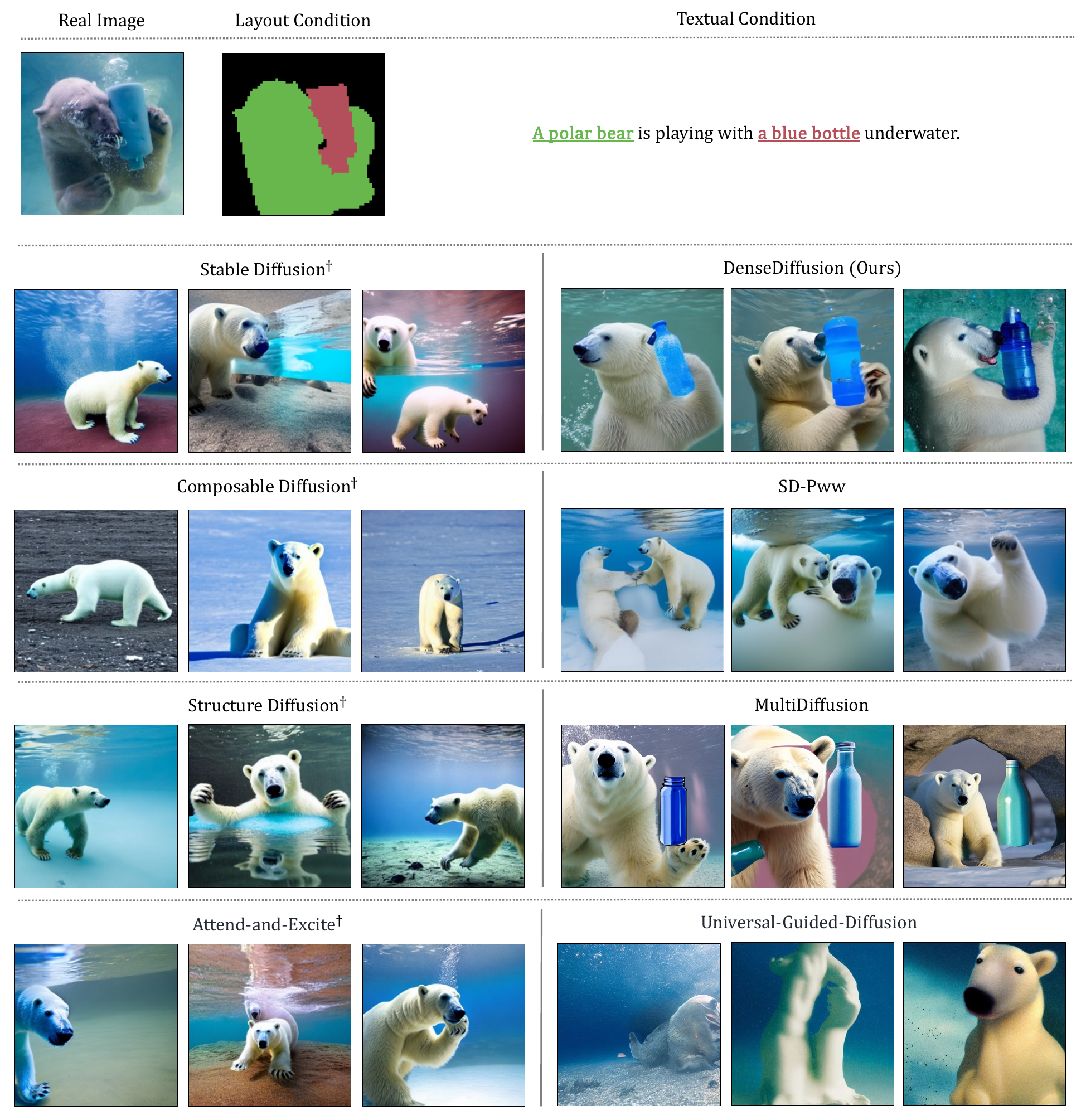}} \\
    \end{tabular}
  \caption{
  Qualitative comparison with additional methods.
  Methods denoted with $^{\dagger}$ receive textual condition only.
  }
  \label{fig:coco_0}
\end{figure*}

\begin{figure*}[t]
    \centering
    \begin{tabular}{c}
    \multicolumn{1}{c}{\includegraphics[width=0.97\linewidth]{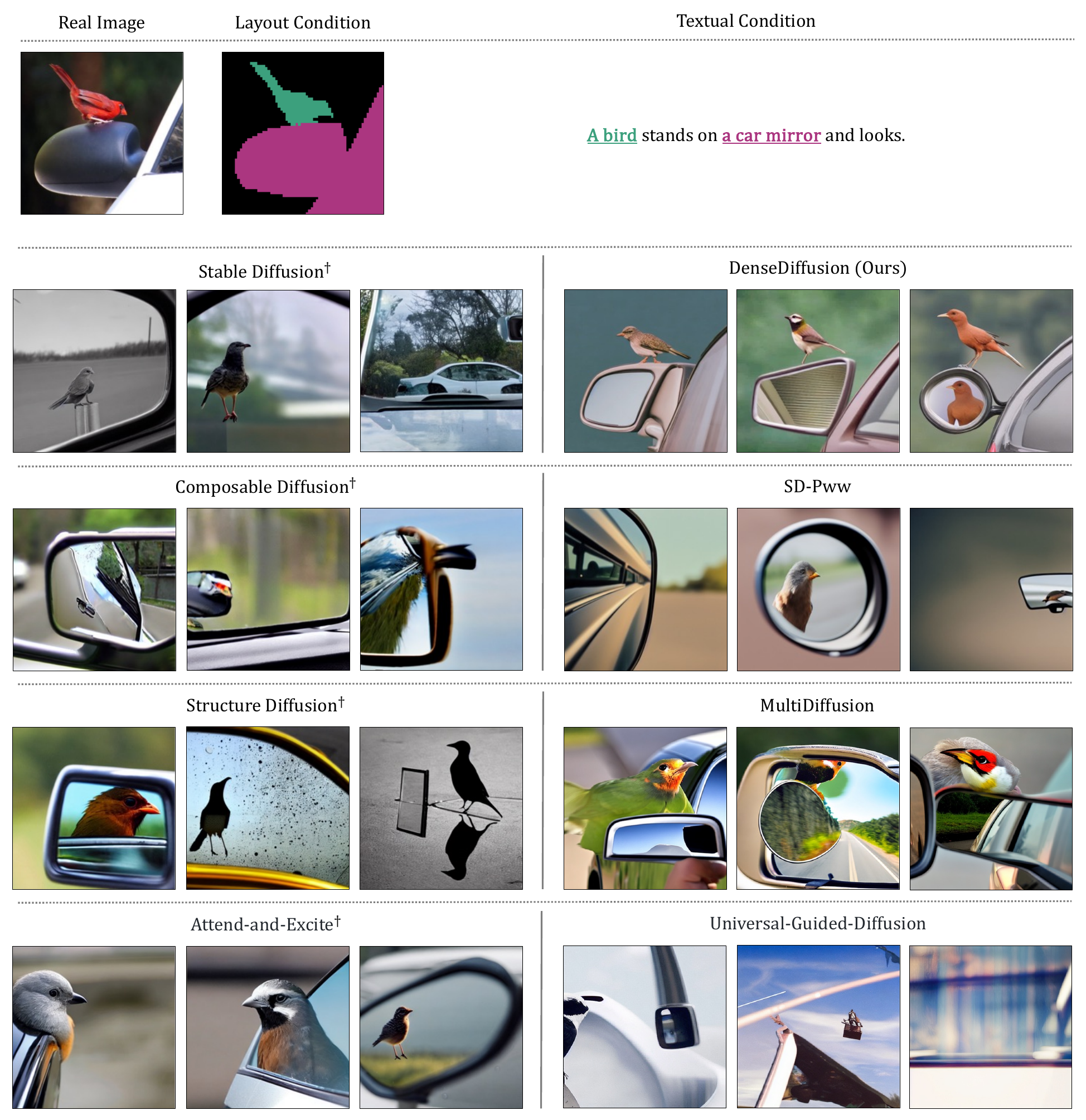}} \\
    \end{tabular}
  \caption{
  Qualitative comparison with additional methods.
  Methods denoted with $^{\dagger}$ receive textual condition only.
  }
  \label{fig:coco_1}
\end{figure*}

\begin{figure*}[t]
    \centering
    \begin{tabular}{c}
    \multicolumn{1}{c}{\includegraphics[width=0.97\linewidth]{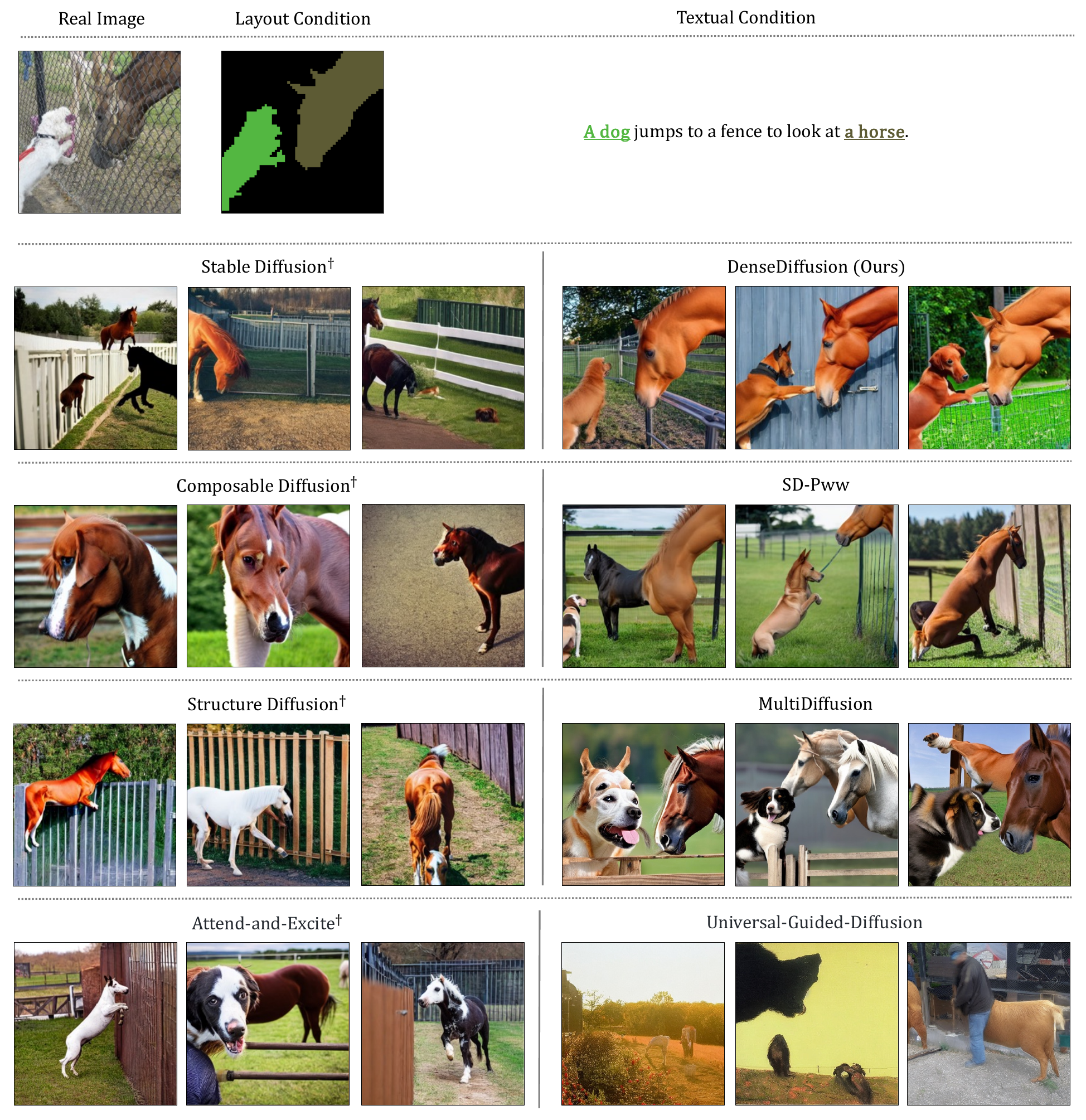}} \\
    \end{tabular}
  \caption{
  Qualitative comparison with additional methods.
  Methods denoted with $^{\dagger}$ receive textual condition only.
  }
  \label{fig:coco_2}
\end{figure*}

Figures~\ref{fig:custom_0},~\ref{fig:custom_1}, and \ref{fig:custom_2} show results for our curated prompts, and Figures~\ref{fig:coco_0},~\ref{fig:coco_1}, and \ref{fig:coco_2} show results for MS-COCO prompts. We observe that our attention modulation method is more effective than  SD-Pww~\cite{ediffi} and Structure Diffusion~\cite{strucdiff} regarding the fidelity to textual and layout conditions. 

MultiDiffusion~\cite{multidiffusion} accurately reflects objects, but the results often look like collages of multiple objects, due to the lack of interaction between independent denoising processes. 
Meanwhile, Composable Diffusion~\cite{compdiff}, Attend-and-Excite~\cite{attendnexcite} and Universal-Guided-Diffusion~\cite{universaldiffusion} often fail to follow the conditions well and tend to generate unrealistic images.

\end{document}